\pdfoutput=1

\documentclass[11pt]{article}

\usepackage{acl}

\usepackage{times}
\usepackage{latexsym}

\usepackage[T1]{fontenc}

\usepackage[utf8]{inputenc}

\usepackage{microtype}

\usepackage{amsmath}

\usepackage{amsfonts}

\usepackage{soul}
\usepackage{tikz}
\usepackage{longtable}
\usepackage[export]{adjustbox}
\usepackage{subcaption}
\usepackage{makecell}

\usepackage[figuresright]{rotating}

\usepackage{cleveref}
\usepackage{xcolor, color}
\usepackage{graphicx}
\usepackage{multirow}
\usepackage{booktabs}
\usepackage{comment}
\usepackage{float}
\usepackage{enumitem}
\usepackage{ctable} 
\usepackage{anyfontsize} 
\usepackage{array}
\newcolumntype{P}[1]{>{\raggedright\arraybackslash}p{#1}}

\definecolor{lightblue}{RGB}{219,226,238}
\definecolor{darkred}{RGB}{105,17,10}
\definecolor{lightyellow}{RGB}{251,242,214}
\definecolor{darkyellow}{RGB}{82,58,34}
\definecolor{lightgrey}{RGB}{230,230,230}
\definecolor{darkgrey}{RGB}{57,57,57}

\newcommand{\systemname}{{\sc CovidKB~}}

\newcommand{\roundcolor}[2]{\begin{tikzpicture}[baseline=(rel.base)]\node(rel)[draw,rounded corners,fill=#1]{#2};\end{tikzpicture}}

\newcommand{\newparagraph}[1]{\vspace{2pt} \noindent \textbf{#1}}

\usepackage{xurl}

\title{Extracting a Knowledge Base of COVID-19 Events from Social Media}

\author{Shi Zong$^1$\quad Ashutosh Baheti$^2$\quad Wei Xu$^2$\quad Alan Ritter$^2$\\
  $^1$University of Waterloo\quad $^2$Georgia Institute of Technology\\
    {\tt s4zong@uwaterloo.ca, abaheti95@gatech.edu}\\
    {\tt \{wei.xu, alan.ritter\}@cc.gatech.edu}}

\begin{document}

\maketitle

\begin{abstract}

We present a manually annotated corpus of 10,000 tweets containing public reports of five COVID-19 events, including positive and negative tests, deaths, denied access to testing, claimed cures and preventions. 
We designed slot-filling questions for each event type and annotated a total of 28 fine-grained slots, such as the location of events, recent travel, and close contacts. 
We show that our corpus can support fine-tuning BERT-based classifiers to automatically extract publicly reported events, which can be further collected for building a knowledge base.
Our knowledge base is constructed over Twitter data covering two years and currently covers over 4.2M events.
It can answer complex queries with high precision, such as \textit{``Which organizations have employees that tested positive in Philadelphia?''}
We believe our proposed methodology could be quickly applied to develop knowledge bases for new domains in response to an emerging crisis, including natural disasters or future disease outbreaks.\footnote{Our corpus (with user-information removed), automatic extraction models, and the corresponding knowledge base are publicly available at \url{https://github.com/viczong/extract_COVID19_events_from_Twitter}.}
\end{abstract}

\section{Introduction}

\label{sec:intro}

\begin{figure}[!t]
    \centering
    \includegraphics[width=0.45\textwidth, cfbox=gray 1pt 1pt]{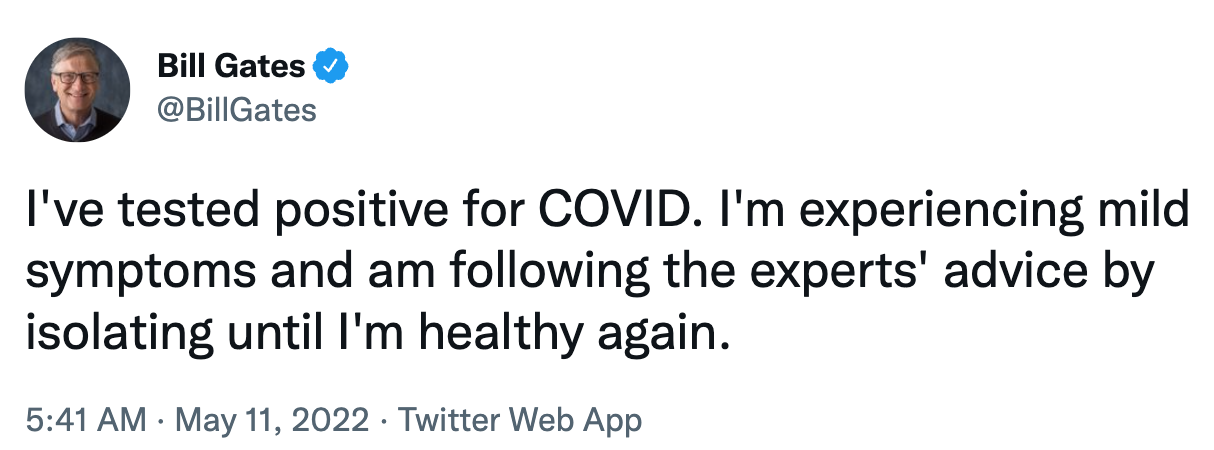}
    \caption{Example tweet that contains a self-reported {\sc Tested Positive} event.} 
    \label{fig:example_tweets}
\end{figure}

\begin{figure*}[!h]
    \centering
    \includegraphics[width=0.92\textwidth]{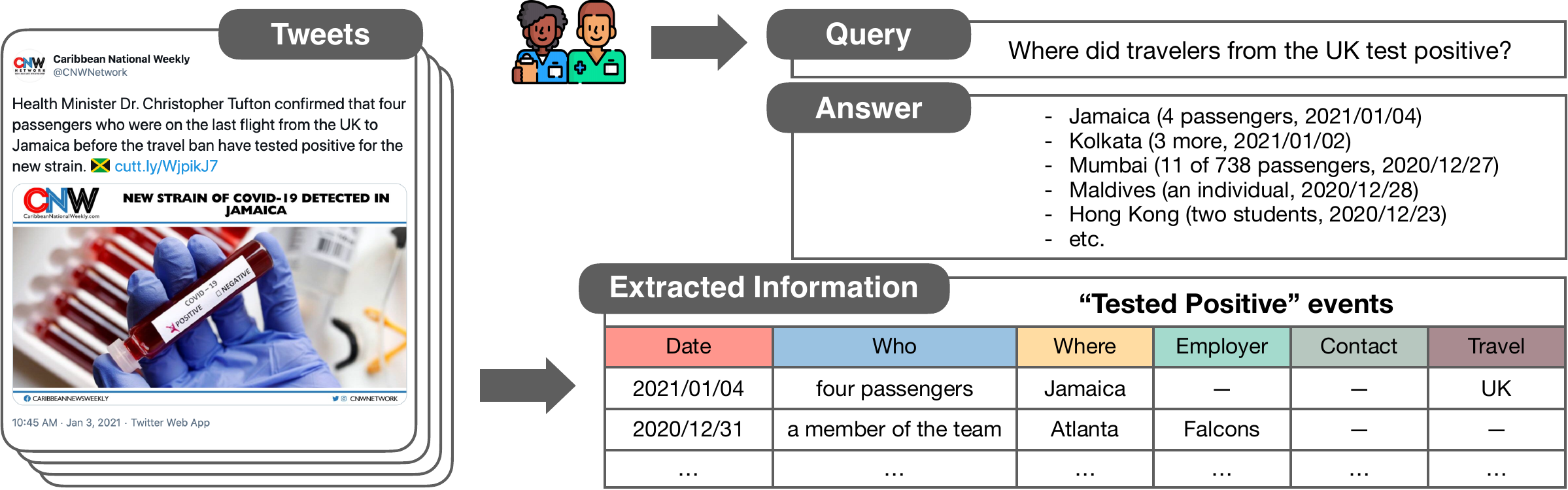}
    \caption{
    Overview of our COVID-19 event extraction system, which continuously extracts and indexes structured information about publicly reported events from Twitter. Users can enter structured queries to retrieve relevant tweets, such as {\tt \{location:?, travel:UK\}} to find test positive cases that are exported from the UK.
    }
    \label{fig:semantic_interface}
\end{figure*}

Since December 2019, the novel coronavirus rapidly spread across the world, and consequently, a flood of COVID-19 related information has appeared on social media. This includes reports on public figures who have tested positive/negative for the virus, which often break first on Twitter, such as Bill Gates's announcement as shown in Figure \ref{fig:example_tweets}. 
Besides public figures, individual users and organizations on Twitter also report COVID-19 events around the world. 
For example in January 2021, many sources in different countries reported an increasing number of new cases exported from the UK (Figure \ref{fig:semantic_interface}).
Being able to gather this information can potentially help experts and the general public to quickly identify issues and assess the situation near real-time, complementing officially reported data which may take longer to obtain, and does not include information at the same level of granularity as that reported in natural language on news and social media.

In this paper, we present an empirical study on the extraction of large quantities of structured knowledge related to an ongoing pandemic from Twitter. To achieve this, we construct a corpus of 10,000 tweets with rich linguistic annotations, covering five event types: positive tests, negative tests, denied access to testing, deaths, claimed methods of cure and prevention. More specifically, we annotate fine-grained semantic information for each event type by designing slot-filling questions and asking annotators to highlight text spans as answers. We show that our corpus can support training BERT-based classifiers to extract structured information automatically from Twitter. While slot F1 scores vary from 0.3 to 0.9 in individual tweets (most F1 scores are greater than 0.5), we show it is possible to achieve very high accuracy by aggregating extractions over a large corpus, exploiting redundancy of information that arises when events are widely discussed on Twitter. 
Although many Twitter datasets have emerged after the COVID-19 outbreak, to the best of our knowledge, our work is the first to provide complex linguistic annotations to support structured information extraction.

To demonstrate the utility of our dataset, we built {\sc CovidKB}, a knowledge base that supports structured queries over COVID-19 events, by indexing events extracted by our model over millions of tweets. Our system allows users to execute structured search queries over the extracted events, answering questions such as {\em ``Which organizations in Houston have reports of employees who tested positive?''} or {\em``Who tested positive that had close contact with Boris Johnson?''} (see Figure \ref{fig:semantic_interface}). We envision \systemname could help address the issue of information overload for professionals \cite{zhang2020rapidly} who need to stay on top of recent developments related to COVID-19, including journalists \cite{karmakharm-etal-2019-journalist}, epidemiologists and public policymakers.
Our extractor can also detect claims about methods of cures and prevention of the disease, which could be useful in helping to track online misinformation \cite{thorne-etal-2018-fever,stefanov-etal-2020-predicting,uci_misinfo}.

\section{Related Work}

\newparagraph{Event Extraction from Twitter.} There has been much interest in extracting events from Twitter. For example, \citet{10.1145/2339530.2339704} built a system for open domain event extraction. Recent work also explored extraction of cybersecurity events \cite{10.1145/2736277.2741083,chang2016expectation}, including denial of service attacks \cite{chambers2018detecting} and software vulnerabilities \cite{zong2019analyzing}. \citet{zhou-etal-2017-event} use a nonparametric Bayesian mixture model for event extraction. In this work, we design event types and attributes that are specific for COVID-19 and develop automatic NLP tools for extracting structured information from tweets. 

\newparagraph{Existing COVID-19 Datasets.} 
There have been many datasets that collect tweets related to COVID-19 \citep{info:doi/10.2196/19273, b2020largescale}. However, most are either unlabeled or provided with general-purpose NLP model predictions, rather than structured linguistic annotations of COVID-specific information, as in this work. 
For example, Twitter officially releases a stream with predicted entities 
(such as {\it person} and {\it place}) and topic labels (such as {\it sports} and {\it movies}).
\citet{10.1145/3404820.3404823} released a COVID-19 collection of geo-located tweets that contain COVID relevant keywords and hashtags.
\citet{dimitrov2020tweetscov19} put together 8 million tweets with automatically generated entity linking and sentiment scores.
\citet{hu-etal-2020-weibo} presented a large-scale dataset of 40 million raw posts from Weibo with no annotations.
There also exist a few datasets that contain human annotations at the time of writing. For example, \citet{uci_misinfo} annotated 5,000 tweets for studying COVID-19 misconceptions. \citet{covid19tweet} classified 10,000 tweets as informative and uninformative.
\citet{DBLP:journals/corr/abs-2010-03824} annotated a dataset of mechanism relations from COVID-19 related scientific papers.
Compared to prior work, we provide more fine-grained human annotations on text spans with predefined slots for COVID-19 events. Our annotations can support training supervised learning models that are capable of extracting structured information \cite{adrian-bejan-harabagiu-2014-unsupervised,venugopal-etal-2014-relieving}, similar to other influential datasets in information extraction and question answering, such as KBP \cite{tac2011overview} and SQuAD \cite{rajpurkar-etal-2016-squad}.


\newparagraph{Social Media Monitoring for Public Health.}
Analyzing social media and other user-generated web data for monitoring public health has been an active research area. 
For example, Google Flu Trends (GFT) uses search engine query data to detect influenza epidemics \cite{Ginsberg:2009ug}. \citet{paul2014twitter} use the Twitter message content to forecast influenza rates.  
GFT has been found to over-estimate influenza-like illness \cite{lazer2014parable}.  In contrast to GFT, our main focus is to develop methods that process large quantities of raw tweets into a \textit{structured} format to help people find specific information, rather than forecasting or nowcasting official statistics.


\section{An Annotated Corpus for COVID-19 Event Extraction}
\label{sec:corpus}

To extract structured knowledge from tweets, we formulate the problem as a supervised slot filling task \citep{jurafskyspeech, benson2011event, tac2011overview}.
Specifically, given a tweet, annotators are asked to first identify whether it contains a relevant event, then highlight the text spans of answers that correspond to a list of pre-defined questions for each event type (detailed questions are in \Cref{tb:questions}).

\subsection{Data Collection}
\label{sec:data_collection}

We consider five event types related to COVID: \textsc{Tested Positive}, \textsc{Tested Negative}, \textsc{Can Not Test}, \textsc{Death}, and \textsc{Cure \& Prevention}. The design of these event types is inspired by the statistics reported in Johns Hopkins COVID-19 dashboard, which are of interest to the public and epidemiologists.\footnote{ \url{https://coronavirus.jhu.edu/map.html}}
The first four types aim to extract structured information about events related to COVID-19, many of which are news stories about public figures. 
We have been continuously collecting Twitter data related to COVID-19 since 2020/01/15 by tracking relevant keywords using the Twitter API, such as \textit{tested positive} for \textsc{Tested Positive} events (see \Cref{tb:data_collection} for a full list of our carefully selected keywords). As we will shown in \Cref{sec:dynamics}, our fixed set of keywords are able to track the evolution of pandemic even over a period of two years, although a dynamic selection of keywords is promising to explore in future work.

\newparagraph{Preprocessing.} 
In this work, we mainly focus on English tweets, identified by using {\ttfamily langid.py} \citep{lui-baldwin-2012-langid}.
We remove retweets and other duplicates, keeping the tweet that was posted earliest. Before de-duplication process, all URLs and user mentions are removed.
We also use Jaccard similarity with a threshold of 0.7 to remove near-identical tweets that are posted same-day.

\begin{table}[!t]
\centering
\resizebox{0.47\textwidth}{!}{
\begin{tabular}{r|ccc}
\toprule
\textbf{Event Type} & \textbf{\# Anno. Total} & \textbf{\# Event Specific} & \textbf{\# Slots}\\\midrule
{\sc Tested Positive} & 3,000 & 2,146 & 9 \\
{\sc Tested Negative} & 1,700 & 893 & 8 \\
{\sc Can Not Test} & 1,700 & 680 & 5 \\
{\sc Death} & 1,800 & 626 & 6 \\
{\sc Cure \& Prev.} & 1,800 & 832 & 3\\
\midrule
\textbf{Total} & 10,000 & 5,177 & 31 \\
\bottomrule
\end{tabular}
}
\caption{Statistics of COVID-19 Twitter Event Corpus.}
\label{tb:dataset_statistics}
\end{table}

\subsection{Annotation Process}
\label{sec:annotation}

We randomly sample 10,000 tweets from five event types to annotate. The train and dev sets consist of 7,500 annotated tweets, that were published between 2020/01/15 and 2020/04/26. To construct the test set, we annotated 2,500 tweets, 500 for each event type, that were published from a later time period between 2020/04/27 and 2020/06/27.
This simulates a real-world scenario that a model is trained on historical records and then applied to future data. \Cref{tb:dataset_statistics} shows the overall statistics of our labeled corpus.

\subsubsection{Two-phase Annotation}

Given a tweet, annotators are asked to first identify whether it contains a relevant event, then highlight the text spans of answers that correspond to a list of pre-defined questions for each event type in Table \ref{tb:questions}.
We hire crowd workers on Amazon's Mechanical Turk to annotate our full dataset.
Each of the 10,000 tweets is annotated by 7 crowd workers in two steps. 
We paid crowd workers \$0.4-0.5 per HIT and gave extra bonuses to annotators with high annotation quality. The hourly pay was approximately \$8.55.
The main portion of our annotation interface is shown in \Cref{fig:example_interface}.

\newparagraph{Part 1: Event Specificity.} 
Although tweets have been filtered by keywords for each event type, many of them are generic news reports, such as, \textit{``37\% of those tested under 17 for Coronavirus in California tested positive''}. Since we are interested in capturing tweets with detailed information, we first ask the annotators to judge whether a tweet refers to a specific event. For example, for tweets about positive tests, we ask the annotators whether a tweet is about an individual or a small group of people testing positive. Annotators proceed to the next step only if they answer yes to this question. 

\newparagraph{Part 2: Slot Filling.} 
In the second step, we ask a set of pre-defined questions specifically designed for each event type, as listed in Table \ref{tb:questions}. The annotators are provided with candidate answers, which include all noun phrases and named entities extracted by a Twitter-specific NLP tool \citep{ritter-etal-2011-named},\footnote{ \url{github.com/aritter/twitter_nlp}} in a drop-down list. We also combine noun phrases if they are adjacent or separated by a preposition.\footnote{We notice in some cases these noun phrases are not perfect and may include extra words. Annotators are instructed that a candidate answer should only be chosen when it contains no more than three extra words.} 
We include \textit{author of the tweet} as an additional option for the WHO questions.\footnote{These annotations are used to develop classifiers that can detect and remove instances where users publicly report information about themselves.}
For each tweet, annotators have an average of 10 to 11 possible answers to choose from, and are allowed to choose more than one answer for WH-questions. 

\subsubsection{Inter-annotator Agreement}

During annotation, we track crowd workers' performance by comparing their annotations with the majority vote of other workers and remove workers' qualifications if their F1 scores fall below 0.65.\footnote{For more discussions on managing workers on Amazon Mechanical Turk, we recommend reading: { \url{https://homes.cs.washington.edu/~msap/notes/turking-tips.html}}.} 
For the first step of annotation on specificity, the inter-annotator agreement between crowdsourcing workers is 0.68, measured by Fleiss $\kappa$ \cite{doi:10.1162_coli.07-034-R2}. 
We observe a 0.62 F1 score for selected text spans between annotators in our slot filling task, by using each Turker's annotation in turn as the prediction, and then compare it against answers from all other workers. Same method to calculate inter-annotator agreement for text spans has been used in \citet{yang-etal-2018-hotpotqa} and \citet{ 10.1145/3331184.3331352}.

To further validate the quality of slot-filling annotations from the crowdsourcing workers, we hired an experienced in-house annotator to carefully re-annotate the test set (2,500 tweets total, with 500 from each event; see \Cref{sec:data_collection} for details).
The in-house annotator is paid \$15 per hour.
By comparing crowdsourcing workers with our in-house annotator, we find individual annotators do miss some examples, which is 
similar to previous reports on linguistic annotations on relations and events, such as ACE 2005 \cite{min-grishman-2012-compensating}.
However, by aggregating annotations from multiple crowdsourcing workers,\footnote{We consider to include a span annotation for slot-filling task if 3 out of 7 MTurk annotators agree.} we observe high agreement (an average of 0.72 F1 score) with our in-house annotator. We also ask the in-house annotator to examine a sample of tweets to find answer spans that are not identified as candidates by the automatic NLP tool. We find this scenario occurs in less than 2\% of tweets in our dataset.

\subsection{Corpus Analysis}

\newparagraph{Basic Statistics.}
Our annotated tweets have an average length of 34.6 tokens with a standard deviation of 15.6 tokens. 
We note 41.42\% of the tweets have external links and 29.64\% include hashtags.
Examples of our annotated tweets are in \Cref{tb:anno_exp}.

\newparagraph{Bots and Organizational Accounts.} 
Among all the 9,656 unique users, 2.4\% are potentially bots, as identified by the Botometer API \citep{ICWSM1715587}.
We also note that 4.1\% of tweets about {\sc Cure \& Prevention} are potentially posted by bots. Estimated by the Humanizr \citep{McCorriston2015Organizations}, 18.5\% of user accounts in our data belong to organizations, rather than individuals.

\section{Automatic Event Extraction}
\label{sec:model}

We now use our annotated corpus to train and evaluate supervised learning methods for automatic COVID-19 event extraction. Each slot filling question is treated as a binary classification task: given a tweet $t$ and the candidate span $c$, the classification model $f_{e,s}(t,c) \rightarrow \{0,1\}$ predicts whether $c$ correctly answers the question for the slot $s$ of event type $e$.

\subsection{Experimental Settings}

\newparagraph{Baselines.}
We conduct experiments with two methods for automatic COVID-19 event extraction: 

(1) \textit{Logistic Regression.} We implemented a basic logistic regression classifier using bag-of-ngram features ($n$ = 1, 2, 3). The target chunk $c$ is replaced with a special token before computing $n$-grams.

(2) \textit{Fine-tuning BERT.} We also fine-tune a BERT based classifier  \cite{devlin-etal-2019-bert} that takes a tweet $t$ as input and encloses the candidate phrase $c$ in the tweet with a pair of special entity start \texttt{<E>} and end \texttt{</E>} markers. The BERT hidden representation of token \texttt{<E>} is then fed as input to a linear layer to produce the binary prediction. Since our dataset consists of COVID-19 related tweets, we use COVID-Twitter-BERT \cite[CT-BERT;][]{muller2020covid}, an uncased BERT$_\text{large}$ model pre-trained on 22.5M in-domain tweets, related to COVID-19 (0.6B tokens).

\newparagraph{Implementation Details.} 
By design, many slots within an event are semantically related. For example, the {\tt age} slot is directly related to the {\tt who} slot. During development, we found it beneficial to train the final linear layers of all slots for a given event using the shared CT-BERT parameters.
All shared CT-BERT models are fine-tuned with a 2e-5 learning rate using Adam \cite{kingma2014adam} for 4 epochs. This model has about 345M parameters. 

\subsection{Results}

We evaluate our model performance for event type identification and slot filling on the test data, which consists of 2,500 tweets.
Event types can be directly derived from the slot-filling predictions: an event is identified if text spans are extracted for any of the pre-defined slots associated with the event types by our models.
\Cref{tb:event_type_pred} presents F1 scores on classifying event specific tweets on the test set.
\Cref{tb:results} presents slot filling results of the Logistic Regression, BERT$_\text{large}$ and CT-BERT models, as measured by precision, recall and F1 metrics.\footnote{We omit reporting results for a few slots with less than 20 annotations in test set, such as the {\tt duration} slot for {\sc Tested Negative} and the {\tt when} slot for {\sc Can Not Test}.} 

We observe that CT-BERT gives the best overall performance, which outperforms the bag-of-ngrams baseline. CT-BERT has F1 scores ranging from 0.3 to 0.9, depending on the slot for extracting events from individual tweets. 
The F1 score for most slots is greater than 0.5 and the final micro average F1 achieved by CT-BERT is 0.67.
While we do notice some slots have low F1 scores, these slots are normally associated with few annotations in the train set.
Besides, we will show in \Cref{sec:semantic_search} that the performance of our CT-BERT model is sufficient to support the development of a knowledge base, which achieves much higher accuracy for COVID-19 event extraction from Twitter by aggregating extractions over a large volume of tweets.

\begin{table}[h!]
\centering
\small
\begin{tabular}{r|ccc}
\toprule
\textbf{Event Type} & \textbf{BERT} & \textbf{CT-BERT} \\\midrule
{\sc Tested Positive}  & 0.90 & 0.89\\
{\sc Tested Negative} & 0.72 & 0.77\\
{\sc Can Not Test} & 0.72 & 0.73 \\
{\sc Death} & 0.73 & 0.79\\
{\sc Cure \& Prevention} & 0.64 & 0.70\\
\bottomrule
\end{tabular}
\caption{F1 scores for classifying event specific tweets.}
\label{tb:event_type_pred}
\end{table}

\scriptsize
\begin{table}[h!]
\centering
\resizebox{0.95\linewidth}{!}{
\begin{tabular}{l|c|c|c|ccc}
\toprule
\multicolumn{2}{c|}{\sc Tested Positive} & \multicolumn{1}{c|}{Logistic} & \multicolumn{1}{c|}{BERT} & \multicolumn{3}{c}{CT-BERT} \\ 
\multicolumn{1}{l|}{\centering \textbf{Slot}} & \multicolumn{1}{c|}{\textbf{\#}} & \multicolumn{1}{c|}{\textbf{F1}} & \multicolumn{1}{c|}{\textbf{F1}} & \multicolumn{1}{c}{\textbf{P}} & \multicolumn{1}{c}{\textbf{R}} & \multicolumn{1}{c}{\textbf{F1}} \\
\midrule
who & 375 & .48 & .82 & .86 & .82 & \textbf{.84}\\
close contact & 61 & .02 & .44 & .65 & .61 & \textbf{.63}\\
relation & 21 & 0.0 & .51 & .83 & .48 & \textbf{.61}\\
employer & 121 & .15 & .44 & .65 & .54 & \textbf{.59}\\
recent travel & 27 & 0.0 & \textbf{.36} & .44 & .26 & .33 \\
when & 22 & .05 & .38 & .47 & .36 & \textbf{.41}\\
where & 176 & .27 & .60 & .91 & .49 & \textbf{.64}\\
\bottomrule
\toprule
\multicolumn{2}{c|}{\sc Tested Negative} & \multicolumn{1}{c|}{Logistic} & \multicolumn{1}{c|}{BERT} & \multicolumn{3}{c}{CT-BERT} \\ 
\multicolumn{1}{l|}{\centering \textbf{Slot}} & \multicolumn{1}{c|}{\textbf{\#}} & \multicolumn{1}{c|}{\textbf{F1}} & \multicolumn{1}{c|}{\textbf{F1}} & \multicolumn{1}{c}{\textbf{P}} & \multicolumn{1}{c}{\textbf{R}} & \multicolumn{1}{c}{\textbf{F1}} \\
\midrule
who & 274 & .23 & .67 & .78 & .68 & \textbf{.73}\\
close contact & 27 & 0.0 & 0.0 & .24 & .48 & \textbf{.32}\\
relation & 56 & 0.0 & \textbf{.55} & .77 & .41 & .53 \\
where & 49 & 0.0 & \textbf{.44} & .36 & .55 & \textbf{.44}\\
when & 27 & 0.0 & 0.0 & .35 & .41 & \textbf{.38}\\
\bottomrule
\toprule
\multicolumn{2}{c|}{\sc Can Not Test} & \multicolumn{1}{c|}{Logistic} & \multicolumn{1}{c|}{BERT} & \multicolumn{3}{c}{CT-BERT} \\ 
\multicolumn{1}{l|}{\centering \textbf{Slot}} & \multicolumn{1}{c|}{\textbf{\#}} & \multicolumn{1}{c|}{\textbf{F1}} & \multicolumn{1}{c|}{\textbf{F1}} & \multicolumn{1}{c}{\textbf{P}} & \multicolumn{1}{c}{\textbf{R}} & \multicolumn{1}{c}{\textbf{F1}} \\
\midrule
who & 153 & .16 & .57 & .77 & .58 & \textbf{.66}\\
relation & 70 & .08 & .37 & .69 & .34 & \textbf{.46}\\
symptoms & 52 & .06 & .43 & .55 & .62 & \textbf{.58}\\
where & 30 & .20 & .44 & .55 & .40 & \textbf{.46}\\
\bottomrule
\toprule
\multicolumn{2}{c|}{\sc Death} & \multicolumn{1}{c|}{Logistic} & \multicolumn{1}{c|}{BERT} & \multicolumn{3}{c}{CT-BERT} \\ 
\multicolumn{1}{l|}{\centering \textbf{Slot}} & \multicolumn{1}{c|}{\textbf{\#}} & \multicolumn{1}{c|}{\textbf{F1}} & \multicolumn{1}{c|}{\textbf{F1}} & \multicolumn{1}{c}{\textbf{P}} & \multicolumn{1}{c}{\textbf{R}} & \multicolumn{1}{c}{\textbf{F1}} \\
\midrule
who & 139 & .29 & .68 & .83 & .76 & \textbf{.79}\\
relation & 37 & 0.0 & .59 & .96 & .65 & \textbf{.77}\\
when & 33 & .26 & \textbf{.75} & .66 & .82 & .73\\
where & 65 & .22 & .54 & .70 & .60 & \textbf{.64}\\
age & 33 & .18 & .78 & .89 & .94 & \textbf{.91} \\
\bottomrule
\toprule
\multicolumn{2}{c|}{\sc Cure \& Prevention} & \multicolumn{1}{c|}{Logistic} & \multicolumn{1}{c|}{BERT} & \multicolumn{3}{c}{CT-BERT} \\ 
\multicolumn{1}{l|}{\centering \textbf{Slot}} & \multicolumn{1}{c|}{\textbf{\#}} & \multicolumn{1}{c|}{\textbf{F1}} & \multicolumn{1}{c|}{\textbf{F1}} & \multicolumn{1}{c}{\textbf{P}} & \multicolumn{1}{c}{\textbf{R}} & \multicolumn{1}{c}{\textbf{F1}} \\
\midrule
opinion & 152 & .08 & .66 & .85 & .59 & \textbf{.69}\\
what & 261 & .22 & .66 & .83 & .64 & \textbf{.72}\\
who & 235 & .08 & \textbf{.51} & .87 & .37 & \textbf{.51}\\
\bottomrule
\toprule
\multicolumn{2}{c|}{\textbf{Micro Average F1}} & \multicolumn{1}{c|}{.25} & \multicolumn{1}{c|}{.62} & \multicolumn{3}{c}{\textbf{.67}}\\
\bottomrule
\end{tabular}
}
\caption{Slot-filling results on the test set for logistic regression, BERT$_\text{large}$ and CT-BERT based classifiers. \# is the count of gold annotations in the test data for each slot type. F1 in bold are highest in their row.}
\label{tb:results}
\end{table}

\normalsize

\section{\systemname Knowledge Base}
\label{sec:semantic_search}

We have built models that can extract structured information related to COVID-19 from individual tweets.
To demonstrate the utility of our annotated dataset and models, we create a knowledge base (Figure \ref{fig:semantic_interface}) that enables structured search over COVID-19 events that are automatically extracted from Twitter.

\subsection{\systemname Overview}
\label{sec:kb_overview}

\newparagraph{\systemname Statistics.}
Until 2022/04/01 (start dates are in \Cref{tb:dataset_statistics}), our \systemname knowledge base has contained around 4.2M extracted events from over 20M raw tweets and is continuously growing by processing tweets daily.
Events are extracted from deduplicated tweets, which follow the same pre-processing steps in \Cref{sec:data_collection}.
Breakdowns of our extracted events are listed in \Cref{tb:kb_breakdown}.

\begin{table*}[h!]
\centering
\small
\begin{tabular}{l|cccc}
\toprule
\textbf{Simple Queries} & \textbf{P@10} & \textbf{P@20} & \textbf{P@50} & \textbf{P@100}\\\midrule
(S-1) Who tested positive on 2021/06/15? & 100 & 100 & 100 & 99\\
(S-2) Who is promoting cures or preventions? & 90 & 90 & 96 & 91 \\
(S-3) Where were people not able to access testing? & 100 & 100 & 100 & 100 \\
(S-4) How long did people wait for negative test results?  & 100 & 85 & 82 & 82 \\
(S-5) Which organizations have employees who tested positive? & 90 & 90 & 90 & 94\\\toprule\toprule
\textbf{Advanced Queries} & \textbf{P@5} & \textbf{P@10} & \textbf{P@20} & \textbf{P@50}\\\midrule
(A-1) Who tested positive that had close contact with Boris Johnson? & 80 & 70 & 60 & 58 \\
(A-2) Who tested positive that has a recent travel to Japan? & 100 &  100 & 100 & 96 \\
(A-3) What methods of cure and prevention do people think are effective? & 80 & 90 & 85 & 88\\
(A-4) Where did people test positive who traveled from the UK? & 100 & 100 & 100 & 100\\
(A-5) Which organizations have employees that tested positive in San Francisco? & 100 & 100 & 90 & 92\\
\bottomrule
\end{tabular}
\caption{Queries used to evaluate results returned by our knowledge base, reported using Precision@$K$. The queries are presented here in natural language for improved readability. Simple queries can be realized as a single {\tt GroupBy} operation; advanced queries contain both {\tt GroupBy} and {\tt Select}. 
For example, the structured query for A-1 is {\tt \{who:?, contact:`Boris Johnson'\}}. 
All queries use the default time range (from 2020/01/15 to 2022/03/01) unless explicitly specified.
}
\label{tb:query}
\end{table*}

\newparagraph{Interacting with {\sc CovidKB}.}
\systemname supports a simple structured query interface where a user specifies one or more text-filters as a query (see \Cref{fig:query_interface}). This includes two SQL operators, {\tt Select} and {\tt GroupBy}. For the event slot queried by the user, using a special token ``{\tt ?}'', our system returns a list of all unique answers, which were extracted from tweets that match the search criteria and sorted by mention frequency. For example, a user might enter the query {\tt \{employer:?, location:`San Francisco'\}}, and the system will return a list of organizations located in San Francisco where one or more employees tested positive. This simple interface enables a rich set of informative queries over events that were automatically extracted by our classification models.

\begin{table*}[!h]
\small
\centering
\renewcommand{\arraystretch}{0.7}
\begin{tabular}{P{0.96\textwidth}}
\toprule
\textbf{(S-1) Who tested positive on 2021/06/15?}\\\midrule
{\fontsize{7.5}{8}\selectfont {\it \sethlcolor{lightblue}\textbf{\textcolor{blue}{\hl{Teofimo Lopez}}} tests positive for COVID-19, entire Triller PPV card pushed back to August (by @mookiealexander) https://t.co/DoaHNb9Z4T}}\\\cmidrule{1-1}
{\fontsize{7.5}{8}\selectfont {\it \sethlcolor{lightblue}\textbf{\textcolor{blue}{\hl{Vaccinated Hawaiian resident}}} tests positive for Delta coronavirus variant https://t.co/0IJ8QfpYS9}} \\\cmidrule{1-1}
{\fontsize{7.5}{8}\selectfont {\it Royal Caribbean cruise ship launch, sailings postponed after \sethlcolor{lightblue}\textbf{\textcolor{blue}{\hl{crew members}}} test positive for COVID-19… https://t.co/VVrOdS6uEX}}
\\
\specialrule{0.1em}{.2em}{.3em} 
{\fontsize{8}{9.5}\selectfont \textbf{(A-1) Who tested positive that had close contact with Boris Johnson?}}\\\midrule
{\fontsize{7.5}{8}\selectfont {\it \#news PM Boris Johnson in self-isolation after coming into contact with \sethlcolor{lightblue}\textbf{\textcolor{blue}{\hl{a lawmaker}}} who tested positive for COVID-19 https://t.co/Kcy2X3M6vJ}}\\\cmidrule{1-1}
{\fontsize{7.5}{8}\selectfont {\it \sethlcolor{pink}\textbf{\textcolor{darkred}{\hl{Jair Bolsanaro}}} has tested positive for Covid-19. Noval Djokovic and Boris Johnson had it. Life sometimes comes a full circle very fast.}}\\\cmidrule{1-1}
{\fontsize{7.5}{8}\selectfont {\it WH says Trump spoke with Boris Johnson and "wished him a speedy recovery" after \sethlcolor{pink}\textbf{\textcolor{darkred}{\hl{the British PM}}} tested positive for coronavirus.}}\\\cmidrule{1-1}
{\fontsize{7.5}{8}\selectfont {\it Boris Johnson's senior adviser, \sethlcolor{lightyellow}\textbf{\textcolor{darkyellow}{\hl{Dominic Cummings}}}, is self-isolating at home after developing \#coronavirus symptoms. http://bbc.in/2WQhbsZ  Last week, the PM and Health Secretary Matt Hancock both tested positive for \#Covid19. WATCH: https://bbc.in/2Jv55xj  \#Newsnight}}\\
\specialrule{0.1em}{.2em}{.3em} 
{\fontsize{8}{9.5}\selectfont \textbf{(A-3) What methods of cure and prevention do people think are effective?}}\\\midrule
{\fontsize{7.5}{8}\selectfont {\it Very good indeed but you need also to remind them keeping \sethlcolor{lightblue}\textbf{\textcolor{blue}{\hl{social distancing}}}, another basic protective measure to prevent the spread of \#covid19.}}\\\cmidrule{1-1}
{\fontsize{7.5}{8}\selectfont {\it Just like washing \sethlcolor{lightgrey}\textcolor{darkgrey}{\textbf{\hl{your hands}}} is necessary to prevent from Coronavirus, inspecting your personal protective equipment https://t.co/xjY7FRgsV1}}\\\cmidrule{1-1}
{\fontsize{7.5}{8}\selectfont {\it Two men in \sethlcolor{lightgrey}\textcolor{darkgrey}{\textbf{\hl{Georgia drank disinfectants}}} in efforts to prevent COVID-19, officials say http://a.msn.com/01/en-us/BB13kJMw?ocid=st…}}\\
\bottomrule
\end{tabular}
\caption{Examples of correct extractions and errors returned by our knowledge base for sample queries. We use different colors for marking the types of extracted text spans (see \Cref{sec:error_analysis} for more details for the error types):
\sethlcolor{lightblue}\textbf{\color{blue}{\hl{correct extraction}}},
\sethlcolor{pink}\textbf{\color{darkred}{\hl{classification errors}}}, \sethlcolor{lightgrey}\color{darkgrey}{\textbf{\hl{segmentation errors}}},
and \sethlcolor{lightyellow}\textbf{\color{darkyellow}{\hl{ambiguous cases}}}.}
\label{tb:errors}
\end{table*}

Table \ref{tb:query} shows a list of example queries supported by {\sc CovidKB}.
Queries are randomly generated by the authors of this paper.
Note that throughout this paper we present queries to our system using natural language questions for the sake of readability.  In each case, translation to a structured query is straightforward. The user specifies zero or more fields to filter on ({\tt Select}) and a single field to group the results by ({\tt GroupBy}). 
As our knowledge base is continuously updating, users can further combine above structured queries with different time ranges (e.g., query S-1 in \Cref{tb:query} sets the start and end dates as 2021/06/15).
We do not address the problem of automatically mapping natural language questions to structured queries \cite{suhr-etal-2018-learning} in this work, though there is significant prior work on this topic \cite{artzi2011bootstrapping,berant2013semantic}.

\subsection{\systemname Evaluation}

\newparagraph{Precision of Top Extractions.}
We evaluate the accuracy of answers returned by our knowledge base using 10 sample queries and manually inspect the correctness of the top $K$ extractions, sorted by frequency (tweets have been deduplicated as mentioned in \Cref{sec:kb_overview}). 
As reported in \Cref{tb:query}, our knowledge base has high precision for nearly all queries, including queries involving slots with few annotations. 
For example, the {\tt duration} slot is excluded in \Cref{tb:results}, because there are fewer than 20 instances in the test set, whereas \systemname still achieves good performance on queries involving this slot, thanks to the redundancy of information in Twitter. \Cref{tb:errors} present outputs returned by our knowledge base.

\newparagraph{Extracted Answer Types.} 
In \Cref{tb:breakdown_types}, we also show a manual analysis of the types of answers, which are correctly extracted by our system for queries that target the {\tt who} slot. We define two answer types: (1) \textbf{Specific} entities, which are clear referents to people (mostly public figures), such as {\it Boris Johnson} and {\it Dominic Cummings}; (2) \textbf{Generic} entities, which are typically nominal references, such as {\it a woman}.  We observe that the percentage of generic answers varies heavily depending on the query.  For example, query A-1 about people who had close contact with Boris Johnson consists almost entirely of references to specific public figures, whereas A-2, about people who tested positive after traveling from Japan yields only generic references.

\begin{table}[!h]
\centering
\small
\begin{tabular}{c|c|cc}
\toprule
\textbf{Query ID} & \textbf{\# Corr / \# All} & \textbf{Specific} & \textbf{Generic}\\\midrule
S-1 & 99 / 100 & 63.6\%  & 36.4\% \\
S-2 & 91 / 100 & 75.8\% & 24.2\% \\\midrule
A-1 & 29 / 50 & 100.0\% & 0.0\% \\
A-2 & 48 / 50 & 6.2\% & 93.8\% \\
\bottomrule
\end{tabular}
\caption{Analysis of answer types in response to the queries (where applicable) in \Cref{tb:query}. The percentage of generic answers varies significantly. 
}
\label{tb:breakdown_types}
\end{table}

\subsection{Error Analysis}
\label{sec:error_analysis}

We perform an error analysis to understand the types of errors our knowledge base contains. Two authors of this paper carefully conducted manual inspections for all the returned results of our sample queries in \Cref{tb:query}.
67 incorrect extractions were identified in 750 extractions,
which can be grouped into four major categories: classification errors (58.2\%), segmentation errors (37.3\%), ambiguous cases (13.9\%) and others (4.5\%). We present some examples of these errors in \Cref{tb:errors}.

\newparagraph{Classification Errors.} We notice our BERT based model struggles with slots that may involve subtle inferences, such as relation or close contact, although the limited number of annotations for these slots might also be a factor in this type of error. 
For example, in the second tweet of query A-1 in \Cref{tb:errors}, the tweet does not imply that {\it Jair Bolsanaro} was in close contact with {\it Boris Johnson}; in the third tweet of query A-1, the model fails to identify that {\it Boris Johnson} and {\it the British PM} refer to the same person.

\newparagraph{Segmentation Errors.} In some cases the extracted items contain extra tokens because of chunker errors, for example {\it georgia drank disinfectants} was extracted as a cure method. We also notice our choice of only extracting noun phrase chunks does not capture verb phrases for the \textsc{Cure \& Prevention} category. For example, instead of extracting {\it washing your hands} and {\it don't touch your face} as prevention methods, our system only extracts {\it your hands} and {\it your face} (see query A-3 in \Cref{tb:errors}). 

\newparagraph{Ambiguous Cases.} In some cases, it is debatable whether an extraction is correct without additional context. For instance in the last tweet of query A-1 in \Cref{tb:errors}, we do not know if \textit{Dominic Cummings} tested positive, although the tweet seems to indicate that he might have been infected.  We consider the extraction to be an error in this case, since the tweet did not specifically mention that he tested positive.

\section{Case Studies}

\subsection{Correlation with Official Data Sources}
\label{sec:dynamics}

\begin{figure}[!h]
    \centering
    \includegraphics[width=0.495\textwidth]{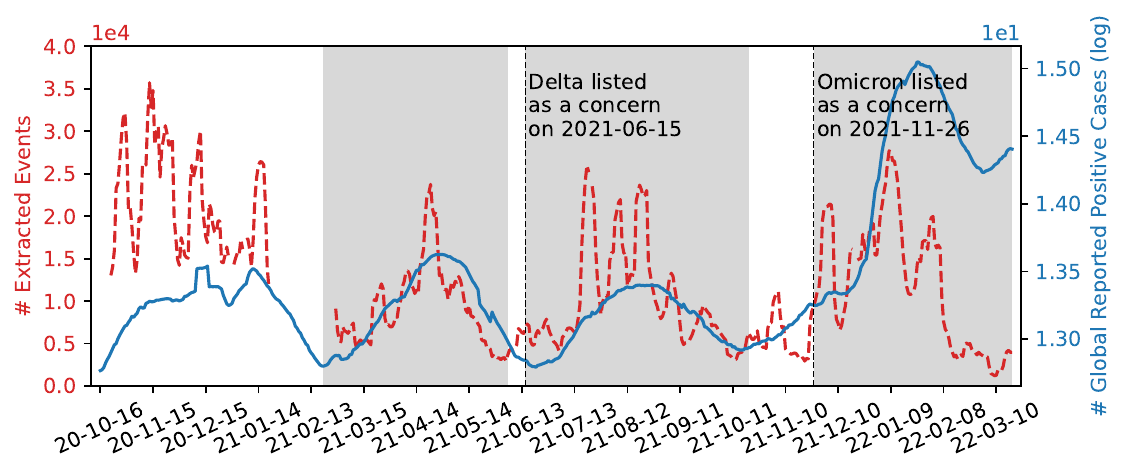}
    \caption{Number of extracted positive events and the actual global reported positive cases (log) show the similar trends in three waves (in grey). Data from 2021/01/21 to 2021/02/26 is missing due to technical issues.}
    \label{fig:actual_positive}
\end{figure}

To investigate whether statistics of events in \systemname correlate with official data sources, we plot the reported global positive cases and the number of extracted tested positive events from our knowledge base over time in \Cref{fig:actual_positive}.
Global reported positive numbers are from Center for Systems Science and Engineering at Johns Hopkins University.\footnote{ \url{https://github.com/CSSEGISandData/COVID-19}}
We use 7-days moving average when drawing two time series curves.
We observe that for both two waves in 2021 and current Omicron wave (highlighted in grey in \Cref{fig:actual_positive}), our extracted events follow similar trend as actual reported cases globally and also show peaks.
This analysis provides evidence to support quality of the extracted information in {\sc CovidKB}, and suggests our knowledge base may contain information that could be used to analyze emerging dynamics of the pandemic. However as mentioned previously, the main use-case for \systemname is to enable semantic search to help journalists, epidemiologists or other professionals quickly analyze information posted on social media.

\vspace{.5cm}

\subsection{Analyzing Claimed Cures and Preventions}

\newparagraph{Public's Attention Shifts over Time.} 
Our knowledge base could also be helpful in monitoring public attention shifts regarding potential treatments and preventative measures over time.
To demonstrate this, we analyze the top frequently mentioned potential cure and prevention methods that people believe are effective within different time ranges (a visualization of top 15 results are in \Cref{tb:sample_sys_output}). Time ranges are roughly divided to follow the global trends of the pandemic shown in \Cref{fig:actual_positive}.

We observe people's opinions regarding certain cure and prevention methods remain unchanged throughout the whole pandemic, including {\it social distancing},  {\it hydroxychloroquine}, {\it (wash) your hands} and {\it masks}.  As time proceeds, there is more focus on medical treatments. For example, \textit{vaccine} and \textit{vaccination} are more frequently discussed. Drugs also draw attention, especially in the last time range (from 2021/10/16 until now): we notice a variety of drugs appear in our knowledge base, including \textit{fluvoxamine}, \textit{monoclonal antibodies}, \textit{AstraZeneca antibody drug} and \textit{Israeli drug}.

We note not all above methods are actually effective for coronavirus. Researchers hold a mixed view for treatments such as {\it hydroxychloroquine}
and {\it ivermectin}.\footnote{For example, Ivermectin has been used in clinical trials: { \url{https://www.covid19treatmentguidelines.nih.gov/therapies/antiviral-therapy/ivermectin/}}. 
However, it is not approved or authorized by FDA: { \url{https://www.fda.gov/consumers/consumer-updates/why-you-should-not-use-ivermectin-treat-or-prevent-covid-19}}.}
This type of automatically extracted information in \systemname could be helpful to track the spread of misinformation online.

\newparagraph{Who is promoting cures?}
We also analyze the returned results from query S-2 to understand who is promoting cures.
A variety of people and organizations are observed, most frequent 10 of which are \textit{Donald Trump}, \textit{China}, \textit{scientists}, \textit{CDC}, \textit{White House}, \textit{Jim Bakker}, \textit{Pfizer}, \textit{Madagascar}, \textit{Dr. Fauci}, and \textit{Bill Gates}.

\section{Conclusion}

In this paper, we presented a corpus of 10,000 tweets annotated with 5 types of events and 28 slots.  We showed that our corpus supports automatic extraction of COVID-19 events using supervised learning.  By aggregating extractions over millions of tweets, our approach can accurately answer a range of structured queries about events that are publicly reported in real-time on Twitter.
Our knowledge base could be a useful tool for epidemiologists, journalists and policymakers to more efficiently track the spread of this new disease.  This work also presents a case-study on how an information extraction system can be rapidly developed for a new domain in response to an emerging crisis.  For example, our methodology could be applied to develop knowledge bases for natural disasters \citep{maza2020event} or future disease outbreaks.

\section*{Ethical Considerations}
\label{sec:discussion}

This study was conducted under the approval of the Institutional Review Board (IRB) of our university and complies with Twitter’s terms of service. 
Following Twitter's policy for content redistribution, we will only release our annotated corpus that contains Tweet IDs (not Tweet Objects) and a list of character offsets corresponding to the annotated mentions. We will not release any user information or demographic data.
Our event extractors produce structured representations of information that was explicitly and publicly stated. We do not derive or infer any potentially sensitive characteristics or health information that may violate users' privacy. Almost all events that are currently indexed by our knowledge base come from public news reports. To further protect users' privacy, we specifically designed two slot-filling questions during annotation in order to detect and remove cases where users publicly report information about themselves, or a person with whom they have a close relationship.

Our knowledge base should be used with caution, as we note the Twitter users are not representative samples of the total population; posts from Twitter users are also not necessarily representative samples of public opinions \cite{sizingup}. 
As Twitter Stream API provides only 1\% of all public tweets, our knowledge base naturally is not able to index all reported cases online. 
Our extractors may contain other unknown biases due to data collection process, for example they might perform worse on African American English.
All these limitations should be taken into consideration in any application that makes use of our data.

\section*{Acknowledgements}

We would like to thank the anonymous reviewers for their valuable suggestions.
This material is based in part on research sponsored by the NSF (IIS-1845670) and IARPA via the BETTER program (2019-19051600004), DARPA via the ARO (W911NF-17-C-0095) in addition to an Amazon Research Award. The views and conclusions contained herein are those of the authors and should not be interpreted as necessarily representing the official policies, either expressed or implied, of IARPA, or the U.S. Government. The U.S. Government is authorized to reproduce and distribute reprints for government purposes notwithstanding any copyright annotation therein.

\bibliographystyle{acl_natbib}
\bibliography{ref}

\begin{thebibliography}{43}
\expandafter\ifx\csname natexlab\endcsname\relax\def\natexlab#1{#1}\fi

\bibitem[{Adrian~Bejan and
  Harabagiu(2014)}]{adrian-bejan-harabagiu-2014-unsupervised}
Cosmin Adrian~Bejan and Sanda Harabagiu. 2014.
\newblock Unsupervised event coreference resolution.
\newblock \emph{Computational Linguistics}.

\bibitem[{Amini et~al.(2021)Amini, Hope, Wadden, van Zuylen, Horvitz, Schwartz,
  and Hajishirzi}]{DBLP:journals/corr/abs-2010-03824}
Aida Amini, Tom Hope, David Wadden, Madeleine van Zuylen, Eric Horvitz, Roy
  Schwartz, and Hannaneh Hajishirzi. 2021.
\newblock Extracting a knowledge base of mechanisms from {COVID-19} papers.
\newblock In \emph{Proceedings of the 2021 Conference of the North American
  Chapter of the Association for Computational Linguistics: Human Language
  Technologies}.

\bibitem[{Artstein and Poesio(2008)}]{doi:10.1162_coli.07-034-R2}
Ron Artstein and Massimo Poesio. 2008.
\newblock Inter-coder agreement for computational linguistics.
\newblock \emph{Computational Linguistics}.

\bibitem[{Artzi and Zettlemoyer(2011)}]{artzi2011bootstrapping}
Yoav Artzi and Luke Zettlemoyer. 2011.
\newblock Bootstrapping semantic parsers from conversations.
\newblock In \emph{Proceedings of the 2011 Conference on Empirical Methods in
  Natural Language Processing}, pages 421--432.

\bibitem[{Banda et~al.(2020)Banda, Tekumalla, Wang, Yu, Liu, Ding, and
  Chowell}]{b2020largescale}
Juan~M. Banda, Ramya Tekumalla, Guanyu Wang, Jingyuan Yu, Tuo Liu, Yuning Ding,
  and Gerardo Chowell. 2020.
\newblock A large-scale {COVID-19} {T}witter chatter dataset for open
  scientific research -- an international collaboration.
\newblock \emph{arXiv preprint arXiv:2004.03688}.

\bibitem[{Benson et~al.(2011)Benson, Haghighi, and Barzilay}]{benson2011event}
Edward Benson, Aria Haghighi, and Regina Barzilay. 2011.
\newblock Event discovery in social media feeds.
\newblock In \emph{Proceedings of the 49th Annual Meeting of the Association
  for Computational Linguistics: Human Language Technologies}.

\bibitem[{Berant et~al.(2013)Berant, Chou, Frostig, and
  Liang}]{berant2013semantic}
Jonathan Berant, Andrew Chou, Roy Frostig, and Percy Liang. 2013.
\newblock Semantic parsing on freebase from question-answer pairs.
\newblock In \emph{Proceedings of the 2013 conference on empirical methods in
  natural language processing}, pages 1533--1544.

\bibitem[{Chambers et~al.(2018)Chambers, Fry, and
  McMasters}]{chambers2018detecting}
Nathanael Chambers, Ben Fry, and James McMasters. 2018.
\newblock Detecting denial-of-service attacks from social media text: Applying
  {NLP} to computer security.
\newblock In \emph{Proceedings of the 2018 Conference of the North American
  Chapter of the Association for Computational Linguistics: Human Language
  Technologies}.

\bibitem[{Chang et~al.(2016)Chang, Teng, and Zhang}]{chang2016expectation}
Ching~Yun Chang, Zhiyang Teng, and Yue Zhang. 2016.
\newblock Expectation-regulated neural model for event mention extraction.
\newblock In \emph{Proceedings of the 2016 Conference of the North American
  Chapter of the Association for Computational Linguistics: Human Language
  Technologies}.

\bibitem[{Chen et~al.(2020)Chen, Lerman, and Ferrara}]{info:doi/10.2196/19273}
Emily Chen, Kristina Lerman, and Emilio Ferrara. 2020.
\newblock Tracking social media discourse about the {COVID-19} pandemic:
  Development of a public coronavirus twitter data set.
\newblock \emph{JMIR Public Health Surveill}.

\bibitem[{Devlin et~al.(2019)Devlin, Chang, Lee, and
  Toutanova}]{devlin-etal-2019-bert}
Jacob Devlin, Ming-Wei Chang, Kenton Lee, and Kristina Toutanova. 2019.
\newblock {BERT}: Pre-training of deep bidirectional transformers for language
  understanding.
\newblock In \emph{Proceedings of the 2019 Conference of the North {A}merican
  Chapter of the Association for Computational Linguistics: Human Language
  Technologies}.

\bibitem[{Dimitrov et~al.(2020)Dimitrov, Baran, Fafalios, Yu, Zhu, Zloch, and
  Dietze}]{dimitrov2020tweetscov19}
Dimitar Dimitrov, Erdal Baran, Pavlos Fafalios, Ran Yu, Xiaofei Zhu, Matthäus
  Zloch, and Stefan Dietze. 2020.
\newblock Tweets{COV}19 -- a knowledge base of semantically annotated tweets
  about the {COVID-19} pandemic.
\newblock \emph{arXiv preprint arXiv:2006.14492}.

\bibitem[{Ginsberg et~al.(2009)Ginsberg, Mohebbi, Patel, Brammer, Smolinski,
  and Brilliant}]{Ginsberg:2009ug}
Jeremy Ginsberg, Matthew~H Mohebbi, Rajan~S Patel, Lynnette Brammer, Mark~S
  Smolinski, and Larry Brilliant. 2009.
\newblock {Detecting influenza epidemics using search engine query data}.
\newblock \emph{Nature}, 457(7232):1012--1014.

\bibitem[{Hossain et~al.(2020)Hossain, Logan~IV, Ugarte, Matsubara, Young, and
  Singh}]{uci_misinfo}
Tamanna Hossain, Robert~L. Logan~IV, Arjuna Ugarte, Yoshitomo Matsubara, Sean
  Young, and Sameer Singh. 2020.
\newblock {COVIDL}ies: Detecting {COVID}-19 misinformation on social media.
\newblock In \emph{Proceedings of the 1st Workshop on {NLP} for {COVID}-19 at
  {EMNLP} 2020}.

\bibitem[{Hu et~al.(2020)Hu, Huang, Chen, and Mao}]{hu-etal-2020-weibo}
Yong Hu, Heyan Huang, Anfan Chen, and Xian-Ling Mao. 2020.
\newblock {W}eibo-{COV}: A large-scale {COVID}-19 social media dataset from
  {W}eibo.
\newblock In \emph{Proceedings of the 1st Workshop on {NLP} for {COVID}-19
  (Part 2) at {EMNLP} 2020}, Online. Association for Computational Linguistics.

\bibitem[{Ji et~al.(2011)Ji, Grishman, and Dang}]{tac2011overview}
Heng Ji, Ralph Grishman, and Hoa Dang. 2011.
\newblock Overview of the {TAC2011} knowledge base population track.
\newblock In \emph{TAC 2011 Proceedings Papers}.

\bibitem[{Jurafsky and Martin(2000)}]{jurafskyspeech}
Daniel Jurafsky and James~H Martin. 2000.
\newblock Speech and language processing: An introduction to natural language
  processing, computational linguistics, and speech recognition.

\bibitem[{Karmakharm et~al.(2019)Karmakharm, Aletras, and
  Bontcheva}]{karmakharm-etal-2019-journalist}
Twin Karmakharm, Nikolaos Aletras, and Kalina Bontcheva. 2019.
\newblock Journalist-in-the-loop: Continuous learning as a service for rumour
  analysis.
\newblock In \emph{Proceedings of the 2019 Conference on Empirical Methods in
  Natural Language Processing and the 9th International Joint Conference on
  Natural Language Processing: System Demonstrations}.

\bibitem[{Kingma and Ba(2015)}]{kingma2014adam}
Diederick~P Kingma and Jimmy Ba. 2015.
\newblock Adam: A method for stochastic optimization.
\newblock In \emph{International Conference on Learning Representations}.

\bibitem[{Lazer et~al.(2014)Lazer, Kennedy, King, and
  Vespignani}]{lazer2014parable}
David Lazer, Ryan Kennedy, Gary King, and Alessandro Vespignani. 2014.
\newblock The parable of google flu: traps in big data analysis.
\newblock \emph{Science}, 343(6176):1203--1205.

\bibitem[{Lee and Sun(2019)}]{10.1145/3331184.3331352}
Grace~E. Lee and Aixin Sun. 2019.
\newblock A study on agreement in pico span annotations.
\newblock In \emph{Proceedings of the 42nd International ACM SIGIR Conference
  on Research and Development in Information Retrieval}, SIGIR'19, page
  1149–1152, New York, NY, USA. Association for Computing Machinery.

\bibitem[{Lui and Baldwin(2012)}]{lui-baldwin-2012-langid}
Marco Lui and Timothy Baldwin. 2012.
\newblock langid.py: An off-the-shelf language identification tool.
\newblock In \emph{Proceedings of the {ACL} 2012 System Demonstrations}, pages
  25--30, Jeju Island, Korea. Association for Computational Linguistics.

\bibitem[{McCorriston et~al.(2015)McCorriston, Jurgens, and
  Ruths}]{McCorriston2015Organizations}
James McCorriston, David Jurgens, and Derek Ruths. 2015.
\newblock Organizations are users too: Characterizing and detecting the
  presence of organizations on twitter.
\newblock In \emph{Proceedings of the 9th International AAAI Conference on
  Weblogs and Social Media}.

\bibitem[{Min and Grishman(2012)}]{min-grishman-2012-compensating}
Bonan Min and Ralph Grishman. 2012.
\newblock Compensating for annotation errors in training a relation extractor.
\newblock In \emph{Proceedings of the 13th Conference of the {E}uropean Chapter
  of the Association for Computational Linguistics}.

\bibitem[{M{\"u}ller et~al.(2020)M{\"u}ller, Salath{\'e}, and
  Kummervold}]{muller2020covid}
Martin M{\"u}ller, Marcel Salath{\'e}, and Per~E Kummervold. 2020.
\newblock {COVID}-{T}witter-{BERT}: A natural language processing model to
  analyse {COVID}-19 content on twitter.
\newblock \emph{arXiv preprint arXiv:2005.07503}.

\bibitem[{Nguyen et~al.(2020)Nguyen, Vu, Rahimi, Dao, Nguyen, and
  Doan}]{covid19tweet}
Dat~Quoc Nguyen, Thanh Vu, Afshin Rahimi, Mai~Hoang Dao, Linh~The Nguyen, and
  Long Doan. 2020.
\newblock {WNUT-2020 Task 2: Identification of Informative {COVID-19} English
  Tweets}.
\newblock In \emph{Proceedings of the 6th Workshop on Noisy User-generated
  Text}.

\bibitem[{Paul et~al.(2014)Paul, Dredze, and Broniatowski}]{paul2014twitter}
Michael~J Paul, Mark Dredze, and David Broniatowski. 2014.
\newblock Twitter improves influenza forecasting.
\newblock \emph{PLOS Currents Outbreaks}.

\bibitem[{Qazi et~al.(2020)Qazi, Imran, and Ofli}]{10.1145/3404820.3404823}
Umair Qazi, Muhammad Imran, and Ferda Ofli. 2020.
\newblock Geo{COV}19: A dataset of hundreds of millions of multilingual
  {COVID-19} tweets with location information.
\newblock \emph{SIGSPATIAL Special}.

\bibitem[{Rajpurkar et~al.(2016)Rajpurkar, Zhang, Lopyrev, and
  Liang}]{rajpurkar-etal-2016-squad}
Pranav Rajpurkar, Jian Zhang, Konstantin Lopyrev, and Percy Liang. 2016.
\newblock {SQ}u{AD}: 100,000+ questions for machine comprehension of text.
\newblock In \emph{Proceedings of the 2016 Conference on Empirical Methods in
  Natural Language Processing}.

\bibitem[{Ritter et~al.(2011)Ritter, Clark, {Mausam}, and
  Etzioni}]{ritter-etal-2011-named}
Alan Ritter, Sam Clark, {Mausam}, and Oren Etzioni. 2011.
\newblock Named entity recognition in tweets: An experimental study.
\newblock In \emph{Proceedings of the 2011 Conference on Empirical Methods in
  Natural Language Processing}.

\bibitem[{Ritter et~al.(2012)Ritter, Mausam, Etzioni, and
  Clark}]{10.1145/2339530.2339704}
Alan Ritter, Mausam, Oren Etzioni, and Sam Clark. 2012.
\newblock Open domain event extraction from {T}witter.
\newblock In \emph{Proceedings of the 18th ACM SIGKDD International Conference
  on Knowledge Discovery and Data Mining}.

\bibitem[{Ritter et~al.(2015)Ritter, Wright, Casey, and
  Mitchell}]{10.1145/2736277.2741083}
Alan Ritter, Evan Wright, William Casey, and Tom Mitchell. 2015.
\newblock Weakly supervised extraction of computer security events from
  {T}witter.
\newblock In \emph{Proceedings of the 24th International Conference on World
  Wide Web}.

\bibitem[{Spiliopoulou et~al.(2020)Spiliopoulou, Maza, Hovy, and
  Hauptmann}]{maza2020event}
Evangelia Spiliopoulou, Salvador~Medina Maza, Eduard Hovy, and Alexander~G
  Hauptmann. 2020.
\newblock Event-related bias removal for real-time disaster events.
\newblock In \emph{Findings of the Association for Computational Linguistics:
  EMNLP 2020}, pages 3858--3868.

\bibitem[{Stefanov et~al.(2020)Stefanov, Darwish, Atanasov, and
  Nakov}]{stefanov-etal-2020-predicting}
Peter Stefanov, Kareem Darwish, Atanas Atanasov, and Preslav Nakov. 2020.
\newblock Predicting the topical stance and political leaning of media using
  tweets.
\newblock In \emph{Proceedings of the 58th Annual Meeting of the Association
  for Computational Linguistics}.

\bibitem[{Suhr et~al.(2018)Suhr, Iyer, and Artzi}]{suhr-etal-2018-learning}
Alane Suhr, Srinivasan Iyer, and Yoav Artzi. 2018.
\newblock Learning to map context-dependent sentences to executable formal
  queries.
\newblock In \emph{Proceedings of the 2018 Conference of the North {A}merican
  Chapter of the Association for Computational Linguistics: Human Language
  Technologies}.

\bibitem[{Thorne et~al.(2018)Thorne, Vlachos, Christodoulopoulos, and
  Mittal}]{thorne-etal-2018-fever}
James Thorne, Andreas Vlachos, Christos Christodoulopoulos, and Arpit Mittal.
  2018.
\newblock {FEVER}: a large-scale dataset for {F}act {E}xtraction and
  {VER}ification.
\newblock In \emph{Proceedings of the 2018 Conference of the North {A}merican
  Chapter of the Association for Computational Linguistics: Human Language
  Technologies}.

\bibitem[{Varol et~al.(2017)Varol, Ferrara, Davis, Menczer, and
  Flammini}]{ICWSM1715587}
Onur Varol, Emilio Ferrara, Clayton Davis, Filippo Menczer, and Alessandro
  Flammini. 2017.
\newblock Online human-bot interactions: Detection, estimation, and
  characterization.
\newblock In \emph{Proceedings of the Eleventh International AAAI Conference on
  Web and Social Media}.

\bibitem[{Venugopal et~al.(2014)Venugopal, Chen, Gogate, and
  Ng}]{venugopal-etal-2014-relieving}
Deepak Venugopal, Chen Chen, Vibhav Gogate, and Vincent Ng. 2014.
\newblock Relieving the computational bottleneck: Joint inference for event
  extraction with high-dimensional features.
\newblock In \emph{Proceedings of the 2014 Conference on Empirical Methods in
  Natural Language Processing}.

\bibitem[{Wojcik and Hughes(2019)}]{sizingup}
Stefan Wojcik and Adam Hughes. 2019.
\newblock Sizing up twitter users.
\newblock Technical report, Pew Internet and American Life Project.

\bibitem[{Yang et~al.(2018)Yang, Qi, Zhang, Bengio, Cohen, Salakhutdinov, and
  Manning}]{yang-etal-2018-hotpotqa}
Zhilin Yang, Peng Qi, Saizheng Zhang, Yoshua Bengio, William Cohen, Ruslan
  Salakhutdinov, and Christopher~D. Manning. 2018.
\newblock {H}otpot{QA}: A dataset for diverse, explainable multi-hop question
  answering.
\newblock In \emph{Proceedings of the 2018 Conference on Empirical Methods in
  Natural Language Processing}, pages 2369--2380, Brussels, Belgium.
  Association for Computational Linguistics.

\bibitem[{Zhang et~al.(2020)Zhang, Gupta, Nogueira, Cho, and
  Lin}]{zhang2020rapidly}
Edwin Zhang, Nikhil Gupta, Rodrigo Nogueira, Kyunghyun Cho, and Jimmy Lin.
  2020.
\newblock Rapidly deploying a neural search engine for the {COVID}-19 open
  research dataset: Preliminary thoughts and lessons learned.
\newblock \emph{arXiv preprint arXiv:2004.05125}.

\bibitem[{Zhou et~al.(2017)Zhou, Zhang, and He}]{zhou-etal-2017-event}
Deyu Zhou, Xuan Zhang, and Yulan He. 2017.
\newblock Event extraction from {T}witter using non-parametric {B}ayesian
  mixture model with word embeddings.
\newblock In \emph{Proceedings of the 15th Conference of the {E}uropean Chapter
  of the Association for Computational Linguistics}.

\bibitem[{Zong et~al.(2019)Zong, Ritter, Mueller, and
  Wright}]{zong2019analyzing}
Shi Zong, Alan Ritter, Graham Mueller, and Evan Wright. 2019.
\newblock Analyzing the perceived severity of cybersecurity threats reported on
  social media.
\newblock In \emph{Proceedings of the 2019 Conference of the North American
  Chapter of the Association for Computational Linguistics: Human Language
  Technologies}.

\end{thebibliography}

\clearpage

\appendix

\setcounter{table}{0}
\renewcommand{\thetable}{A\arabic{table}}
\setcounter{figure}{0}
\renewcommand{\thefigure}{A\arabic{figure}}

\onecolumn

\section{Dataset}

\subsection{Keywords for Data Collection} 

We provide the keywords used for collecting data along with starting date in \Cref{tb:data_collection}.
Keywords in our experiments are carefully chosen to both have a wide coverage of tweets with different linguistic phenomena and have a good precision of collecting tweets that are relevant to our tasks.

\begin{table*}[!htbp]
\centering
\resizebox{0.8\textwidth}{!}{
\begin{tabular}{rcl}
\toprule
 \textbf{Event Type} & \textbf{Start From} & \multicolumn{1}{c}{\textbf{Keywords}}\\\midrule
\multirow{1}{*}{{\sc Tested Positive}} & \multirow{1}{*}{2020/01/15} &
(test {\ttfamily OR} tests {\ttfamily OR} tested) positive {\ttfamily AND} {\fontfamily{cmss}\selectfont VIRUS}\\\cmidrule{1-3}
\multirow{1}{*}{{\sc Tested Negative}} & \multirow{1}{*}{2020/02/15} &
(test {\ttfamily OR} tests {\ttfamily OR} tested) negative {\ttfamily AND} {\fontfamily{cmss}\selectfont VIRUS} \\\cmidrule{1-3}
\multirow{4}{*}{{\sc Can Not Test}} & \multirow{4}{*}{2020/01/15} & 
(can't {\ttfamily OR} can not) get (tested {\ttfamily OR} test {\ttfamily OR} tests)\\
& & (can't {\ttfamily OR} can not) be tested\\
& & (couldn't {\ttfamily OR} could not) get (tested {\ttfamily OR} test {\ttfamily OR} tests)\\
& & (couldn't {\ttfamily OR} could not) be tested\\
\cmidrule{1-3}
\multirow{1}{*}{{\sc Death}} &
\multirow{1}{*}{2020/02/15} & 
(died {\ttfamily OR} pass away {\ttfamily OR} passed away) {\ttfamily AND} {\fontfamily{cmss}\selectfont VIRUS} \\\cmidrule{1-3}
\multirow{1}{*}{{\sc Cure \& Prevention}} &
\multirow{1}{*}{2020/03/01} & 
(cure {\ttfamily OR} prevent) {\ttfamily AND} {\fontfamily{cmss}\selectfont VIRUS} \\
\bottomrule
\end{tabular}
}
\caption{Keywords used for each event type. We consider the following variants for {\fontfamily{cmss}\selectfont VIRUS}: {\fontfamily{cmss}\selectfont VIRUS} =  (COVID19 {\ttfamily OR} COVID-19 {\ttfamily OR} corona {\ttfamily OR} coronavirus).}
\label{tb:data_collection}
\end{table*}

\subsection{Data Annotation}

\begin{figure*}[h!]
    \centering
    \includegraphics[width=\textwidth]{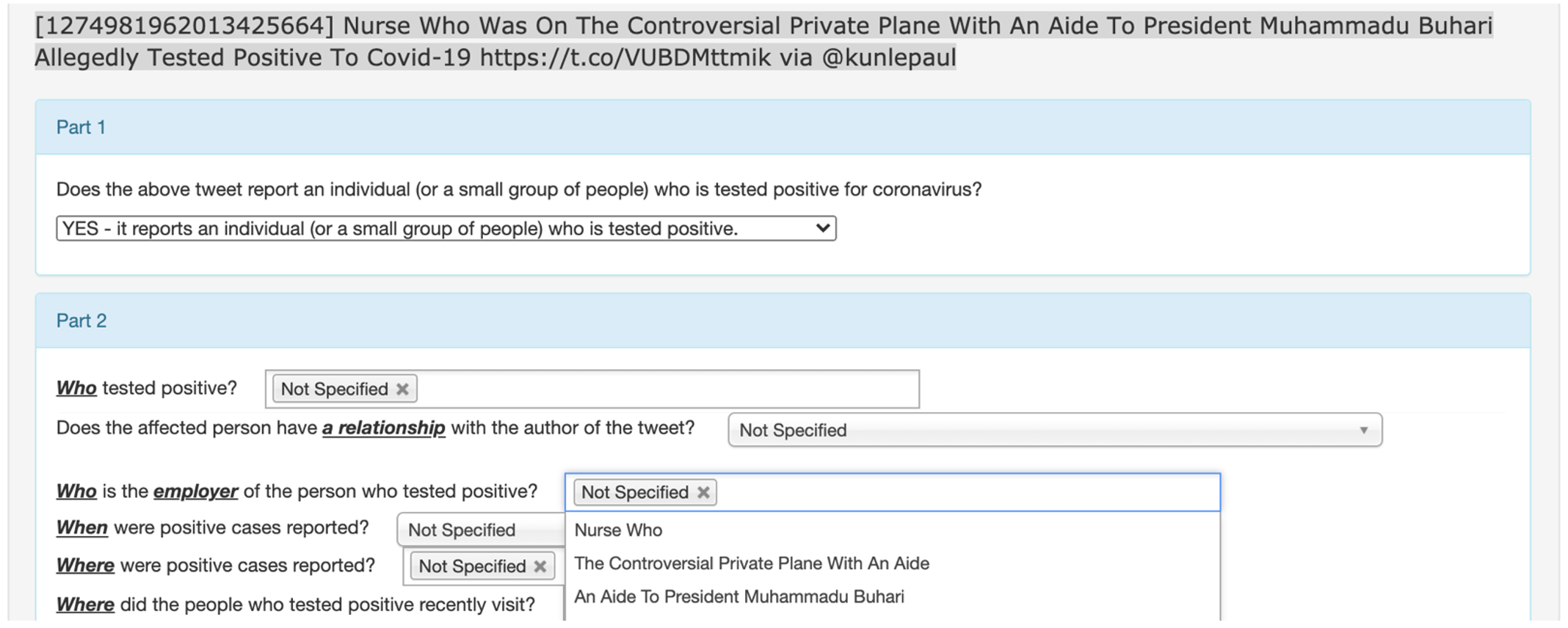}
    \caption{Main portion of the annotation interface shown to Mechanical Turk workers for annotating {\sc Tested Positive} events.}
    \label{fig:example_interface}
\end{figure*}

The complete slot filling questions used for annotating COVID-19 events are listed in \Cref{tb:questions}. We also provide the annotation interface shown to Mechanical Turk workers in \Cref{fig:example_interface}.

\begin{table*}[h!]
\centering
\small
\begin{tabular}{c|c|l}
\toprule
\textbf{Event Type} & \textbf{Slot Name} & \textbf{Slot Filling Questions} \\\midrule
 & who & Who tested positive (negative)? \\
& close contact & Who was in close contact with the person who tested positive (negative)?    \\
{\sc Tested} & relation & Does the affected person have a relationship with the author of the tweet?\\
{\sc Positive} & employer & Who is the employer of the person who tested positive? \\
{------} & recent travel & Where did the people who tested positive recently visit? \\
{\sc Tested} & when & When were positive (negative) cases reported? \\
{\sc Negative} & where & Where were positive (negative) cases reported? \\
& age & What is the age of the people who tested positive (negative)?\\
& duration & How long did it take to know the result of the test? \\ \midrule
& who & Who can not get a test?     \\
\multirow{2}{*}{{\sc Can Not}} & relation & Does the untested person have a relationship with the author of the tweet? \\ 
\multirow{2}{*}{{\sc Test}} & when & When was the person unable to obtain a test? \\
& where & Where was the person unable to obtain a test? \\
& symptoms & Is the affected person currently experiencing any COVID-19 related symptoms? \\\midrule 
\multirow{5}{*}{{\sc Death}}
& who & Who died from COVID-19? \\
& relation & Does the deceased person have a personal relationship with the author of the tweet? \\     
& when & When was the death reported? \\
& where & Where was the death reported? \\
& age & What is the age of the person who died? 
\\\midrule
\multirow{2}{*}{{\sc Cure \&}}& opinion & Does the author of the tweet believe cure/prevention is effective?\\
\multirow{2}{*}{{\sc Prevention}}& what & Which method of cure/prevention is mentioned?\\
& who & Who is promoting the cure or prevention? \\
\bottomrule
\end{tabular}
\caption{Slot filling questions used for annotating COVID-19 events.}
\label{tb:questions}
\end{table*}

\vspace{50pt}

\subsection{Annotated Samples}

Examples of our annotated tweets are presented in \Cref{tb:anno_exp}.

\begin{table*}[h!]
\centering
\small
\resizebox{.99\textwidth}{!}{
\begin{tabular}{r|p{0.78\textwidth}|p{0.18\textwidth}}
\toprule
\textbf{Event Type} & \textbf{Tweet} & \textbf{Annotations} \\ \midrule
{\sc Positive} &
{\it \roundcolor{olive!25}{\#Karnataka} | \roundcolor{blue!30}{\roundcolor{lightgray!50}{A 26-year-old man}} returning from \roundcolor{cyan!50}{\#Greece} tested positive for \#COVID19, becoming the fifth positive case in the state, a health official said on Thursday.  \#CoronavirusPandemic \#COVID \#COVID19india \texttt{[URL]}}
& 
\roundcolor{blue!30}{\textbf{WHO}} \roundcolor{lightgray!50}{\textbf{AGE}} \roundcolor{olive!25}{\textbf{WHERE}} 
\roundcolor{cyan!50}{\textbf{RECENT V.}}
\\\midrule
{\sc Negative} &
{\it Live updates: \roundcolor{blue!30}{Boris Johnson} tested negative for Covid-19 on leaving hospital, says Downing Street  \#coronavirus }
& \roundcolor{blue!30}{\textbf{WHO}} 
\\\midrule
{\sc Death} & {\it ‘\#TopChef Masters' winner Floyd \#Cardoz dies after \#coronavirus diagnosis’  “World-renowned chef \roundcolor{blue!30}{Floyd Cardoz} died \roundcolor{lime!50}{Wednesday} in \roundcolor{olive!25}{New Jersey} at \roundcolor{lightgray!50}{age 59}.”  “Cardoz admitted himself to the hospital on March 17 after feeling feverish.” }
& \roundcolor{blue!30}{\textbf{WHO}} \roundcolor{lightgray!50}{\textbf{AGE}} \roundcolor{olive!25}{\textbf{WHERE}}
\roundcolor{lime!50}{\textbf{WHEN}}\\
\midrule
{\sc Can Not Test} & 
{\it \roundcolor{blue!30}{Nurse working in ITU} couldn’t get tested, \& was told that the test was “very expensive”, so he couldn’t have a test. \texttt{[URL]} …} & \roundcolor{blue!30}{\textbf{WHO}} \\
\bottomrule
\end{tabular}
}
\caption{Examples of our annotated tweets.}
\label{tb:anno_exp}
\end{table*}

\section{\systemname Knowledge Base}

\subsection{Statistics of Our Knowledge Base}

We report the number of extracted events along with the breakdown statistics for each slot in \Cref{tb:kb_breakdown}. 

\begin{table}[h!]
\small
\centering
\resizebox{0.99\textwidth}{!}{
\begin{tabular}{r|r|rrrrrrrrrrrr}
\toprule
\multirow{2}{*}{\textbf{Event Types}} & \multirow{2}{*}{\textbf{\# Extracted}} & \multicolumn{12}{c}{\textbf{Number of Events per Slot}} \\\cmidrule{3-14}
 &  & who & relation & when & where & age & close contact & employer & recent travel & duration & symptoms & opinion & what \\\midrule
{\sc Tested Pos}  & 2,354,363  & 2,098,964  & 164,126 & 81,053 & 602,552  & 32,361 & 122,952 & 264,275 & 84,157 & -- & -- & -- & -- \\\cmidrule{1-14}
{\sc Tested Neg}  & 411,071 &  387,354 & 47,325 & 17,044 & 28,447 & 851 & 7,733 & -- & -- & 9,049 & -- & -- &  -- \\\cmidrule{1-14}
{\sc Can Not Test}  & 30,552 & 26,468 & 17,432 & 94 & 7,637 & -- & -- & -- & -- & -- & 14,881 & -- & -- \\\cmidrule{1-14}
{\sc Death}  & 779,074 & 629,323 & 91,121 & 164,282 & 230,672 & 143,270 & -- & -- & -- & -- & -- & --  & -- \\\cmidrule{1-14}
{\sc Cure \& Prev.} & 665,422  & 319,077 & -- & -- & -- & -- & -- & -- & -- & -- & -- & 270,493 & 461,290 \\\midrule
\textbf{Total} & 4,240,482 & 3,461,186 & 320,004 & 262,473 & 869,308 & 176,482 & 130,685 & 264,275 & 84,157 & 9,049 & 14,881 & 270,493 & 461,290 \\
\bottomrule
\end{tabular}
}
\caption{Number of extracted events, with a breakdown for each slot in our knowledge base. Slot filling questions that are not applied to specific event types are marked with ``--''. }
\label{tb:kb_breakdown}
\end{table}

\clearpage

\subsection{Interface of Our Knowledge Base}

\begin{figure}[!h]
    \centering
    \includegraphics[width=0.65\textwidth, cfbox=gray 1pt 1pt]{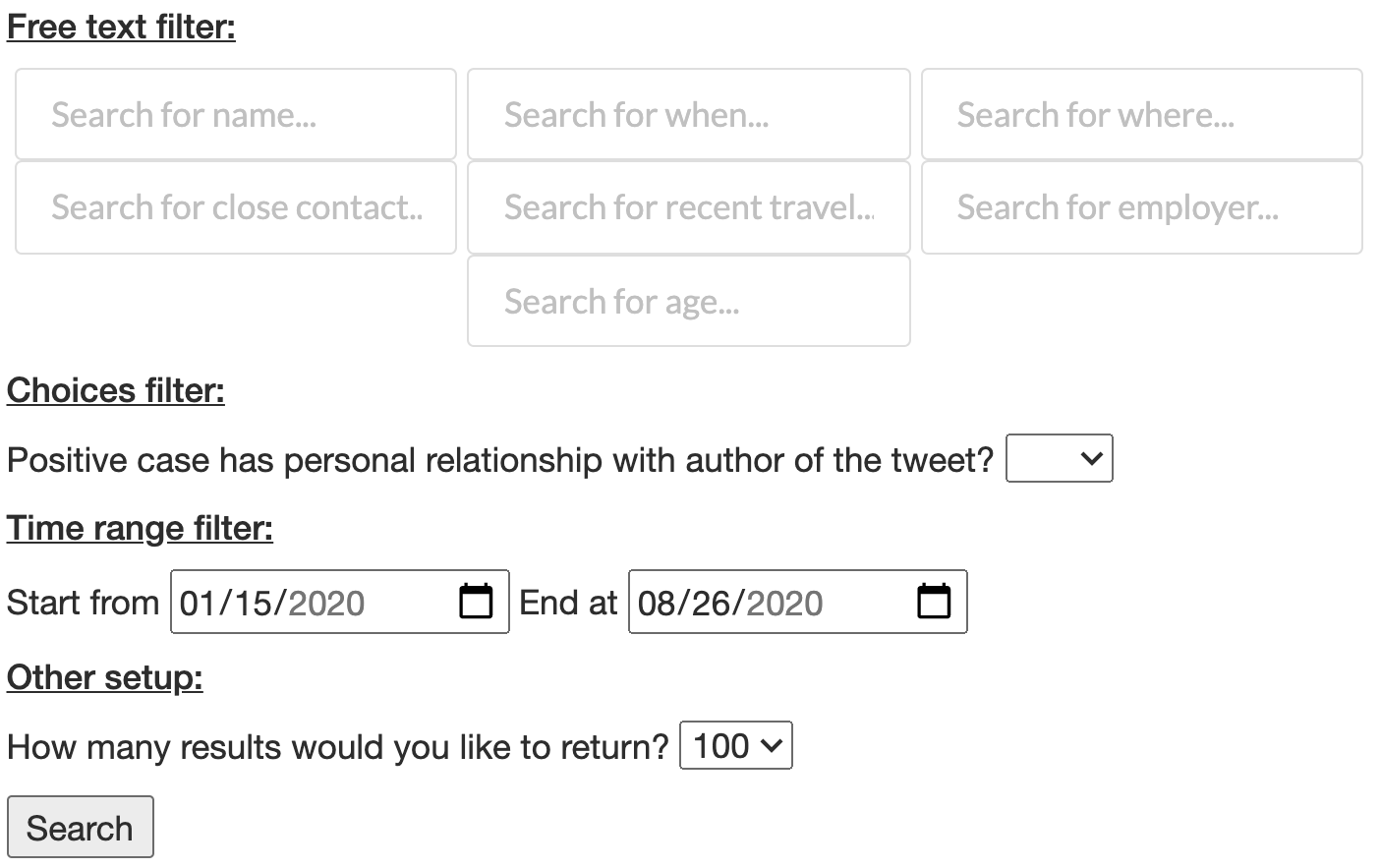}
    \caption{Structured query interface of our knowledge base.}
    \label{fig:query_interface}
\end{figure}

Our structured query interface of the knowledge base is presented in \Cref{fig:query_interface}.

\subsection{Public Attention Shifts for Cure and Prevention Methods over Time}

We present the top 15 frequently mentioned potential cure and prevention methods that
people believe are effective within different time ranges in \Cref{tb:sample_sys_output}. Larger fonts indicate more frequent terms. 

\begin{table*}[h!]
\small
\centering
\begin{tabular}{cc}
\toprule
\multicolumn{2}{l}{\textbf{(A-3) What methods of cure and prevention do people think are effective?}}\\\midrule
\textbf{Before 2021/01/01} & \textbf{From 2021/02/15 to 2021/06/15 (First Wave in 2021)} \\
\includegraphics[width=0.45\textwidth]{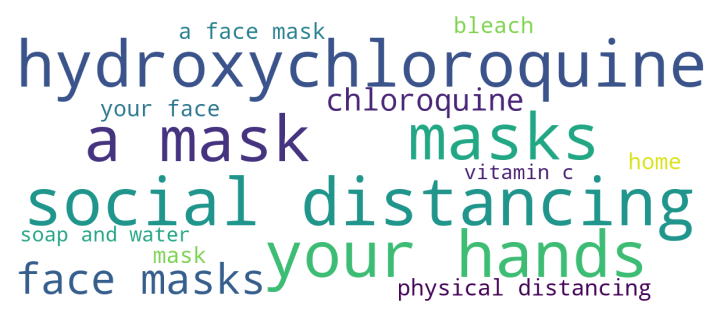} &
\includegraphics[width=0.45\textwidth]{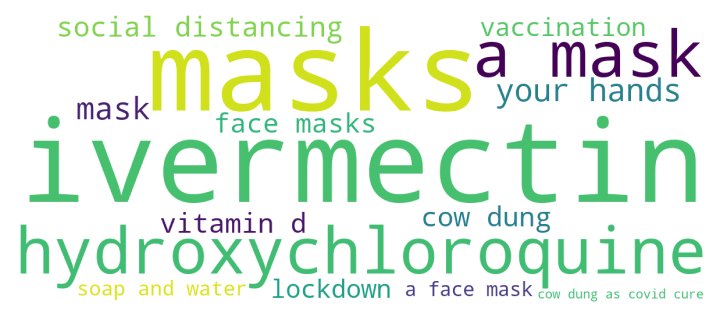} \\\midrule
\textbf{From 2021/06/16 to 2021/10/15 (Second Wave in 2021)} & \textbf{From 2021/10/16 to 2022/04/01} \\
\includegraphics[width=0.45\textwidth]{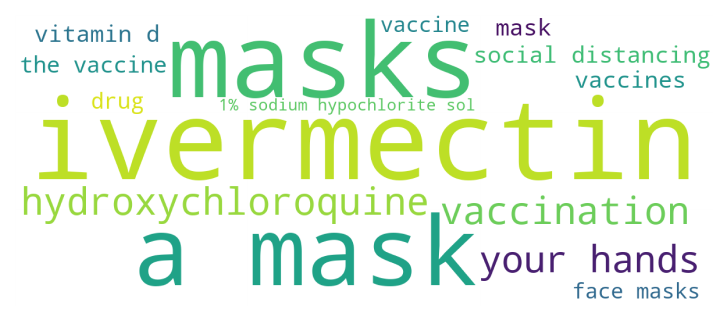} & 
\includegraphics[width=0.45\textwidth]{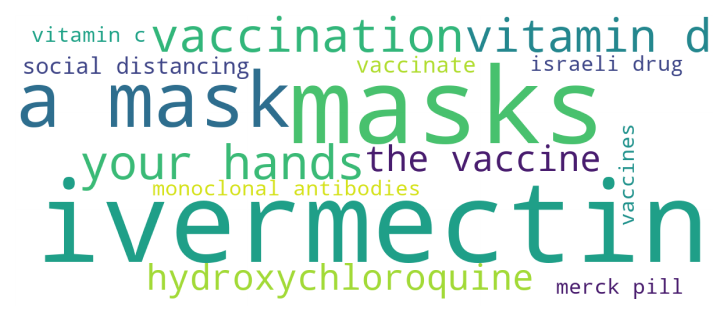} \\
\bottomrule
\end{tabular}
\caption{Top 15 most frequent potential cure and prevention methods that people think are effective over different time ranges.}
\label{tb:sample_sys_output}
\end{table*}

\end{document}